\documentclass{article} 
\usepackage{iclr2020_conference,times}
\usepackage{hyperref}
\usepackage{url}

\usepackage[utf8]{inputenc} 
\usepackage[T1]{fontenc}    
\usepackage{hyperref}       
\usepackage{url}            
\usepackage{booktabs}       
\usepackage{nicefrac}       
\usepackage{microtype}      
\usepackage{framed}
\usepackage{amsmath,amssymb}
\usepackage{placeins}  
\usepackage{color}
\usepackage{xcolor}
\usepackage{framed}
\usepackage{multirow}
\usepackage{graphicx}
\usepackage{verbatim}
\usepackage{wrapfig,lipsum,booktabs}

\usepackage{floatrow}
\newfloatcommand{capbtabbox}{table}[][\FBwidth]

\usepackage{footmisc}  

\usepackage{pifont}
\newcommand{\cmark}{\ding{51}}
\newcommand{\xmark}{\ding{55}}

\newcommand{\ab}[1]{{\color{magenta}[\textbf{AB}:#1]}}

\newcommand{\generategap}[1]{{\textcolor{gray}{[#1]}}}
\newcommand{\markgap}[1]{{\textcolor{red}{[#1]}}}
\newcommand{\markgapsingle}[3]{{\textcolor{red}{[#1]\textsubscript{(#2, #3)}}}}

\title{Real or Fake? Learning to Discriminate Machine from Human Generated Text}

\author{
Anton Bakhtin\thanks{Equal contribution}$^{*\blacklozenge}$ \enskip Sam Gross$^{*\blacklozenge}$ \enskip Myle Ott$^\blacklozenge$ \enskip Yuntian Deng$^\bigstar$ \\
{\bf \enskip Marc'Aurelio Ranzato$^\blacklozenge$ \enskip Arthur Szlam$^\blacklozenge$} \\
$^{\blacklozenge}$ Facebook AI Research \enskip $^\bigstar$ Harvard University \\
  \texttt{\{yolo,sgross,myleott,ranzato,aszlam\}@fb.com} \enskip \texttt{dengyuntian@seas.harvard.edu} \\
}

\iclrfinalcopy 
\begin{document}

\maketitle

\begin{abstract}
Energy-based models (EBMs), a.k.a. un-normalized models, have had recent successes in continuous spaces.
However, they have not been successfully applied to model text sequences.
 While decreasing the energy at training samples is straightforward, mining (negative) samples where
the energy should be increased is difficult.   In part, this is because standard gradient-based methods are not
 readily applicable when the input is high-dimensional and discrete.  Here, we side-step
this issue by generating negatives using pre-trained auto-regressive language models.  The EBM then works
in the {\em residual} of the language model; and is trained to discriminate real text from text generated
by the auto-regressive models.

We  investigate the {\em generalization ability} of residual EBMs, a pre-requisite for using them in
other applications.  We extensively analyze generalization for the task
of classifying whether an input is {\em machine or human} generated, a natural task given the training loss and how
we mine negatives. Overall, we observe that EBMs can generalize remarkably well to changes in the architecture of
the generators producing negatives. However, EBMs exhibit more sensitivity to the training set
used by such generators.
\end{abstract}

\section{Introduction}
Energy-based models (EBMs) have a long history in machine learning~\citep{hopfield, cd, ebm}. Their appeal stems from
 the minimal assumptions they make about the generative process of the data.
Unlike directed or auto-regressive models which are defined in terms of a sequence of conditional distributions,
EBMs are defined in terms of a single scalar energy function, representing the joint compatibility between
all input variables.
EBMs are a strict generalization of probability models, as the  energy function need not be normalized or even have convergent integral.

Training an EBM consists of decreasing the energy function at the observed training data points (a.k.a. positives),
while increasing it at other data points (a.k.a. negatives)~\citep{ebm}.
Different learning strategies mainly differ in how negatives are
mined~\citep{ebm_unsup}.
Some find negatives by gradient descent, or using Monte Carlo methods like Gibbs sampling~\citep{rbm} and hybrid Monte
Carlo~\citep{teh03}, which enable the loss to approximate maximum likelihood training~\citep{cd}.
Other approaches instead use implicit negatives, by
enforcing global constraints on the energy function, like sparsity of the internal representation~\citep{ebm_unsup}, for instance.  GANs \citep{gan} can be interpreted as a particular form of EBM where the negatives are generated by a learned model.

While there are works exploring the use of EBMs for modeling images~\citep{teh03, ranzato13, mordatch19},
they have not been successfully applied to text.
One reason is that text consists of sequences of  discrete variables, which makes the energy function not differentiable with respect to its inputs.
Therefore, it is not possible to mine negatives using gradient-based methods. Other approaches to mine
negatives are also not immediately applicable or may be too inefficient to work at scale.

In this work, we start from the observation that current large auto-regressive locally-normalized language models are already strong ~\citep{gpt2}, and therefore, it may be beneficial to use them to constrain the search space of negatives.
We propose to learn in the {\em residual space} of a pre-trained language model (LM),
which we accomplish by using such LM to generate negatives for the EBM.
Given a dataset of positives and pre-generated negatives, the EBM can be trained
using either a binary cross-entropy loss or a ranking loss, to teach the model to assign a lower energy
to true human generated text than to the text generated by the pre-trained LM.

The question we ask in this work is whether such an EBM can generalize well.  Understanding this is important for two reason. First, this generalization is a prerequisite for using residual EBMs for modeling text. Second, in our setting, this generalization question is equivalent to the question of whether it is possible for a learned model (the energy function) to discriminate real text from text generated by an auto-regressive model.  Discriminating real vs. machine-generated text is an important task on its own that has recently gained a lot of attention~\citep{GLTR, gpt2, zellers19}.

Our contribution is an extensive study of the generalization ability of such residual EBMs, or in other words, the generalization ability of models trained to detect real text from machine generated text.
In particular, we assess how well the energy function is robust to changes in the architecture of the generator
and to changes in the data used to train the generator. The overall finding is that the energy function is remarkably
robust, and the bigger the model and the longer the generation the better its performance.
Moreover, the energy function is robust to changes in the architecture of the
LM producing negatives at test time. However, it is sensitive to the training dataset of the test generator.

\section{Related Work}
Our work can be interpreted as a particular instance of EBMs~\citep{ebm} where negatives
are produced by a pre-trained language model as opposed to the energy function itself.
Learning a generator and a discriminator relates also to Generative Adversarial Networks~\citep{gan}, except that in our case
the generator is trained beforehand.

Since the discriminator is learned after the generator has been trained, it learns from the {\em residual error}
of the generator, and therefore, our training procedure is a particular instance of a ``cascade'' model~\citep{cascade}
and ``boosting''~\citep{boosting}.

Using a separately trained scoring function to evaluate and rank candidate outputs has a long history which dates back to work
on parsing and machine translation~\citep{rerankingMT}. In that work however, the goal was to improve a weak generator
by employing a linear reranker taking as input relatively few hand-design features.
The approach has been recently re-discovered in the context
of dialogue modeling by~\cite{kulikov2018importance}, but here negatives are randomly chosen next utterances from the
training dataset.

Several recent works have studied whether machine generations can be detected automatically,
but they do not study how these findings {\it generalize} to settings where generator architectures
and corpora are different between training and test time. For example, \citet{zellers19} (GROVER) assume that the
generator is known and apply only slight fine-tuning in order to train the energy function.
Similarly, \citet{GLTR} (GLTR)  assume knowledge of the generator; these Authors say
``{\em We further hypothesize that these methods generalize to black-box scenarios,
as long as the fake text follows a similar sampling assumption and is generated by a large language model}'';
our work answers precisely this question, provides a rigorous experimental protocol and quantitative results.

Finally, there has been a release of a training dataset of the GPT-2
language model generations~\citep{gpt2gen}
for the purpose of training discriminators capable of detecting machine generated text. While we share
the same motivation, our work is a much broader investigation on the topic. We assess generalization
of several discriminator architectures to not just one but several kinds of generators and
corpora used for training (including GPT-2).

\section{Energy-Based Models for Text}
In this section, we describe how we train the energy based model and how we mine negatives.

\subsection{Learning} \label{sec:approach}
Our goal is to learn an energy function $E(w_1, \dots,  w_n | c; \theta) \in \mathbb{R}$
that scores the {\em joint compatibility} of an input sequence of tokens $(w_1, \dots,  w_n)$ given some context $c$ and a
set of parameters $\theta$. The context depends on the application, it could be the preceding text, some keywords,
a bag of words, a title, etc. In this work for simplicity, $c$ is an affix from which we condition the generation.

The goal of training is to assign to golden sequences, i.e. sequences taken from a dataset of human generated text,
 lower energy than other sequences. We parameterize the energy function as a neural network,
using the architectures described in \textsection\ref{sec:expscorer}.

At training time, the energy function can be trained using a variety of different losses.
In this work, we consider two choices: the binary cross-entropy loss and
the ranking loss~\citep{collobert2011natural}. As the findings are similar,
unless otherwise specified we will refer to the binary cross-entropy loss, and report results with the ranking
loss in Appendix~\ref{sec:ranking}.

Let $x^+$ be a positive sample taken from the training set, and consisting of a {\em sequence} of $n$
tokens given some context $c$.
Let $(x^-_1, ..., x^-_k)$ be a set of $k$ negative samples each derived from the same context $c$ as above,
all containing at least some machine generated tokens. We train our energy function using the (per-sample)
binary cross-entropy loss:
\begin{equation}
\mathcal{L}_{\mbox{BCE}} = - \log( \sigma(- E(x^+|c;\theta))) + \log( \sigma(- E(\hat{x}^- | c; \theta)))
\label{eq:bceloss}
\end{equation}
where $\hat{x}^-$ is the most offending negative~\citep{ebm}, i.e. its index is the solution of
$\arg \min_{i=1}^k E(\hat{x}^-_i|c;\theta)$, and $\sigma$ is the sigmoid function: $\sigma(u) = \frac{1}{1+\exp(u)}$.

\subsection{Generating Negatives} \label{sec:negatives}
The most critical component of training an energy based model is the method used to generate {\em negatives},
 i.e. inputs where the energy should score high (unlikely inputs).
In settings with continuous variables, researchers have suggested MCMC~\citep{teh03} or Langevin dynamics~\citep{mordatch19}.
In this work instead, we use the fact that modern auto-regressive models for text are already quite good,
and we use them for negative sampling.

We train two auto-regressive language models, a left-to-right one which will be used to produce suffixes assuming the prefix
is the context, and a right-to-left one which will be used to generate prefixes assuming the suffix is the context.
The negatives are generated by top-k sampling~\citep{fan18} setting $k$ equal to 10.
Given a trained language model (for instance, a left-to-right autoregressive model) and given a positive example
$x^+=(w_{i+1}, \dots, w_{i+n})$,
for a given context:
$c=(w_{1}, \dots, w_{i})$,
a negative can be written as:
$x^-=(\hat{w}_{i+1}, \dots, \hat{w}_{i+n})$,
 where $w_j$ for $j \in [1, i+n]$ are ground truth words, the first $i$ of them belonging to the common context,
 and $\hat{w}_{j}$ for $j \in [i+1, i+n]$ are words generated by the language model conditioned on $c$.
In the same way, we can sample a negative with a right-to-left  model yielding $x^-=(\hat{w}_1, \dots, \hat{w}_{n})$,
for a given context $c = (w_{n+1}, \dots, w_{n+i})$.

\section{Experimental Setup} \label{sec:exp_setup}
In this section we first describe the datasets and preprocessing used,
provide architecture details for both generators and scoring functions, and finally
introduce the evaluation settings.

\subsection{Corpora}
\label{sec:data}
We train models on three corpora coming from different domains. We report more detailed statistics about the sizes of these corpora in Appendix Table~\ref{tab:datasets}:
\smallskip
\newline
\noindent
{\bf Books:} The Toronto books corpus described in \citet{Zhu_2015_ICCV, kiros2015skip}, which consists of fiction books in 16 different genres,
totaling about half a billion words.
\smallskip
\newline
\noindent
{\bf CCNews:} We collect a de-duplicated subset of the English portion of the CommonCrawl news dataset \citep{ccnews}, which totals around 16 Billion words.
\smallskip
\newline
\noindent
{\bf Wikitext:}  The wikitext103 dataset from \cite{merity2016pointer}, which consists of 103 million words from English Wikipedia articles.

While Wikitext and CCNews are factual, Books is fiction and comprises a wide variety of writing styles.
The CCNews corpus has the narrowest domain and it is two orders of magnitude larger than Wikipedia.
Overall, these datasets are interesting because they enable us to assess the ability of the energy function
to fit and generalize across various axes, from the amount of data available at training time
to the richness of style and relatedness among the different  data sources.

On Wikitext and Books, we extract positive sequences from windows of text that are 160 tokens long with a stride of 40.
On the larger CCNews we do the same except that we stride by 160 tokens.
This protocol to mine positives is used both at training and test time, although at test time we limit the evaluation to
60,000 randomly chosen positive samples.

We use a Byte Pair Encoding~\citep{sennrich2015neural} in order to represent all the dataset with a common vocabulary.
In particular, our vocabulary contains 50k tokens that was constructed from a byte level UTF-8 encoding of the
CC-News corpus following~\citet{gpt2}.


\subsection{Generator Architectures} \label{sec:generarch}
We mainly use a transformer based network~\citep{vaswani2017attention} to generate negatives.
We have a medium, large and huge transformer model based on the
architecture used in~\citet{baevski2018adaptive}, yielding three language
models in total: TransfSmall, TransfBig and TransfHuge; see details also in Appendix~\ref{sec:sizes}.

The small sized models use 6 blocks each containing a multi-head attention module with 8 heads. The large
models use 12 blocks each containing a multi-head attention module with 16 heads.
The huge models use 48 blocks each containing a multi-head attention module with 25 heads.
Transformer
models are also implemented in~\citet{ott2019fairseq} as "transformer\_lm", "transformer\_lm\_big",
and "transformer\_lm\_gpt2\_big".
The TransfHuge has 10x the number of parameters than TransfBig and it is trained on CCNews only.
For each architecture except for TransfHuge we train two models on each each dataset: left to right and right to left.

In addition to the transformer generator, we also  consider a 12-layer convolutional architecture
(Conv)~\citep{dauphin2017language}, and we also use a the third-party trained GPT2 models~\citep{gpt2} as described
in \textsection\ref{sec:cross-corpus}.

As described in \textsection\ref{sec:negatives}, we use these language models to generate either a prefix or a suffix.
Unless otherwise specified, the context is long either $120$ or $140$ tokens (with equal probability).
Positive and negative examples have $40$ or $20$ tokens depending on the context size, for an overall length of $160$ tokens in
all cases. In preliminary experiments, we found that increasing the size of the generations and reducing the size of the context
makes the learning task significantly easier. We analyze the effect of the context size in \textsection\ref{sec:ablation}.

\subsection{EBM Architectures} \label{sec:expscorer}
We consider three architectures for the energy function:
\smallskip
\newline
\noindent
{\bf Linear} which computes an energy value via a bag of tokens: $f(w_1, ..., w_n)  = \left(\sum_{i=1}^n u_{w_i}\right)$,
where $u_i$ is a learnt scalar parameter corresponding to the $i$-th token in the vocabulary.
\smallskip
\newline
\noindent
{\bf BiLSTM}~\citep{birnn,bilstm} which computes an energy value through $L$ bidirectional layers using LSTM recurrent units~\citep{lstm}, as in
 $\mbox{Linear}(\mbox{AvgPool}(h_{L,1},\dots,h_{L,n}))$, where $h_{L,i}$ is the hidden state at position $i$ and layer $L$ which is the concatenation of the forward and
backward hidden states, AvgPool averages hidden states over positions and Linear is a vector of parameters projecting the hidden state down to a scalar value.
We consider two versions, referred to as ``BiLSTMsmall'' and ``BiLSTMbig''.
Both have 4 layers, but BiLSTMsmall has 512 units in both the embedding layer and the hidden layers, while
BiLSTMbig has 758 units in the embedding layer and 2014 units in the  hidden states.
\smallskip
\newline
\noindent
{\bf Transformer}~\citep{vaswani2017attention,bert} which computes an energy value similarly to the BiLSTM's,
except that each bi-LSTM layer is replaced
by a either a bidirectional Transformer layer (BiTransf), or a Transformer with causal self-attention (UniTransf).
For unidirectional models we use the same averaging technique as with BiLSTM models.
For bidirectional models the energy is computed via:
$f(w_1, ..., w_n)  = u^\top h_{L,1} + b$, where $h_{L,1}$ is the top layer hidden state at the first position (as common practice also in prior work~\citep{bert}).
BiTransf uses the BERT-Large architecture~\citep{bert} initialized from~\citet{liu2019roberta}.
It uses 24 self-attention layers with 1024 units and 16-head attention each.
UniTransf has instead 12 layers with 1024 units and 16 attention heads per layer and it is initialized from a language modeling
task as in~\citet{gpt2}.

For all models, we use Adam~\citep{kingma2014adam} optimizer with warmup.
Training is stopped after processing 2.5M samples without any improvement on the validation set.
We use data-parallel synchronous multi-GPU training with up to 8 nodes, each with 8 Nvidia V100 GPUs.
To improve training speed, we use mixed precision training\footnote{\url{https://github.com/NVIDIA/apex}}.
Following common practice we clip the norm of the gradient vector~\citep{pascanu2013difficulty}.
More details about hyper-parameter setting can be found in Appendix Table~\ref{tbl:hyperparams},
while Table~\ref{tbl:scoring_models_size} in Appendix reports the number of parameters of each energy function.

\subsection{Evaluation} \label{sec:eval}
\begin{table}[t]
\center
\begin{tabular}{lcc}
\toprule
& \textsc{corpus:} & \textsc{generator architecture:} \\
& $C_{\text{train}} = C_{\text{test}}$ & $A_{\text{train}} = A_{\text{test}}$ \\
\midrule
in-domain & \cmark & \cmark \\
cross-architecture & \cmark & \xmark \\
cross-corpus & \xmark & \cmark \\
unseen & \xmark & \xmark \\
\bottomrule
\end{tabular}
\caption{\small Four evaluation settings considered in this work, described in \textsection\ref{sec:eval}.
\label{tab:eval_setup}}
\end{table}

We evaluate the generalization of a residual EBM  in four settings: in-domain, cross-architecture, cross-corpus, and  unseen.

These settings are determined by the corpora $C_{\text{train}}$ used to train the training generator $G_{\text{train}}$ with architecture $A_{\text{train}}$  and the corpora $C_{\text{test}}$ used to train the testing generator $G_{\text{test}}$ with architecture $A_{\text{test}}$. Note that $G_{\text{train}}\neq G_{\text{test}}$ even if $A_{\text{test}}=A_{\text{train}}$  as we use different training seeds.  In all cases, $C_{\text{train}}$ is used for fitting  the $G_{\text{train}}$   and also for the positives for the EBM.




 In the {\bf in-domain} setting,  $C_{\text{test}}$ is $C_{\text{train}}$ (but any affixes used as conditioning during testing  are from the test-set of the corpus), and $A_{\text{test}}=A_{\text{train}}$.
In the {\bf cross-architecture} setting, again $C_{\text{test}}$ is $C_{\text{train}}$, but $A_{\text{test}}$ is different from $A_{\text{train}}$.
  In the {\bf cross-corpus} setting, $A_{\text{test}} =A_{\text{train}}$ but $C_{\text{test}}$ is different than $C_{\text{train}}$, and $G_{\text{test}}$ is trained on the training split of $C_{\text{test}}$, while $G_{\text{train}}$ trained on the train split of $C_{\text{train}}$.  In the {\bf unseen} setting, both
$C_{\text{test}}$ is different than $C_{\text{train}}$ and $A_{\text{test}}$ is different from $A_{\text{train}}$.

In all settings, we report performance in terms of average classification accuracy balancing the positive and negative classes.

\section{Results} \label{sec:results}
We now present the main results of this work and extensively investigate the generalization ability of the energy functions we have considered.
\vspace{-.3cm}
\subsection{In-domain generalization} \label{sec:in_domain}
In Table~\ref{tab:in_domain} we report the results of the in-domain generalization experiment using our large language model, TransfBig.
We observe that when the EBMs have similar representational power compared with the generator
(UniTransf, see Table~\ref{tbl:scoring_models_size}),
they are able to distinguish real from fake completions fairly accurately, reaching an accuracy of more than 90\%
on the Books dataset (which is easier since it exhibits the larger variety of style and topics),
and attaining above 88\% on the more challenging CCNews dataset (for which generation is easier and hence
discrimination harder). The Wikipedia dataset has lower accuracy because the EBM overfits to this smaller dataset.

Weaker energy models are able to do comparably or better at discriminating real from fake than
the training generator used as a discriminator by taking the log probability of the sequence as energy.

\begin{table}[t]
\vspace{-.5cm}
  \center
\begin{tabular}{lrrr}
\toprule
 &  Books &  CCNews &  Wiki \\
\midrule
Linear                    &            59.8 &                 58.6 &            56.3 \\
BiLSTMsmall                    &            84.7 &                 77.6 &            71.0 \\
BiLSTMbig                &            86.7 &                 80.1 &            72.8 \\
UniTransf                 &            91.7 &                 88.4 &            76.4 \\
\midrule
{\em TransfBig (log-likelihood)} &            57.1 &                 50.8 &            50.5 \\
\bottomrule
\end{tabular}
\caption{\small ``In domain'' generalization accuracy of EBMs (each row) on various text corpora.
A column corresponds to the corpus used to get positives and to fit the train and test language models,
which are TransfBig (\textsection\ref{sec:generarch})
with different initial seeds.  The last row is the accuracy when using as energy the log-probability of the
training language model over the whole sequence.
\label{tab:in_domain}}
\end{table}
\vspace{-.3cm}
\subsection{Cross-architecture generalization}  \label{sec:cross-arch}
\begin{table}[t]
\vspace{-0.2cm}
\begin{tabular}{lrr}
\toprule
{} &  Conv &  TransfSmall  \\
\midrule
Conv &                    92.9 &            81.2 \\
TransfSmall         &                    86.5 &            87.9 \\
\bottomrule
\end{tabular}
\caption{\small Cross-architecture generalization accuracy using the Wikitext dataset for both training
and testing ($C_{\text{train}} = C_{\text{test}}$).
Each row is a model architecture used for generating the training negatives ($A_{\text{train}}$), and
each column is a model architecture for generating the testing negatives ($A_{\text{test}}$).
The energy function is UniTransf.}
\label{tab:cross_architecture}
\end{table}
In Table \ref{tab:cross_architecture},
 we assess how well the UniTransf energy function
generalizes to different generator architectures at test time, namely Conv and TransfSmall.
As a reference on the Wikitext dataset,
the test perplexity of Conv and TransfSmall are
 35.4 and 33.5, respectively. Therefore, these two generators attain roughly the same perplexity, despite
Conv having about 4 times more parameters, see Table~\ref{tbl:generation_models_size}.

Surprisingly, UniTransf has significantly harder time discriminating
TransfSmall negatives with an in-domain rate of 87.9\%, compared to 92.9\% of  Conv.
Also, UniTransf  trained with TransfSmall negatives is more robust to the (weaker) Conv generations, than vice versa,
with a mild 1.4\% accuracy drop.
However, if we average
values across rows, we see that UniTransf tested with mixed negatives is just slightly more accurate
when training with the harder negatives produced by TransfSmall.

\vspace{-.3cm}
\subsection{Cross-Corpus generalization}\label{sec:cross-corpus}
In Table~\ref{tab:cross-corpus} we show the results of generalizing across corpora using UniTransf as an energy function
and TransfBig as generator both at training and test time.
We observe that models generalize less well across corpora; 
for instance, when testing on Wikitext an energy function
trained with either Books or CCNews, the accuracy is 59.1\% and 65.5\%, respectively.
However, training on the union of two of the corpora
gives a large benefit over training on just one or the other when testing on the third.

Finally, training on the union of {\em all} the three corpora (last two rows)
yields an energy function that is very robust to the testing conditions, with an accuracy which is on par if not better
than training on in-domain data, even for the largest CC-News dataset (second column).

We also tested the bidirectional transformer energy function BiTransf with 355M parameters (almost twice as UniTransf),
 and found that on CC-News it improves accuracy by more than 5\% when it is trained on the union of all corpora,
confirming the finding that bigger models trained on more data can achieve substantially better discrimination.
As BiTransf was pre-trained using the whole Wikipedia rather than the training part of Wikitext103,
we do not report its accuracy on Wiki test set.

\begin{table*}[t]
\center
\begin{tabular}{lccc}
\toprule
\multirow{2}{*}{\textsc{train corpora}} \quad \, & \multicolumn{3}{c}{\textsc{test corpora}} \\
                &  Books            &  CCNews                &  Wiki             \\
\midrule
Wiki                &            70.9 &                 73.6 &            76.4 \\
\hline
Books               &            91.7 &                 63.5 &            59.1 \\
Books + Wiki        &            91.5 &                 73.6 &            78.3 \\
\hline
CCNews               &            60.6 &                 88.4 &            65.5 \\
Books + CCNews       &            90.4 &                 88.5 &            68.3 \\
CCNews + Wiki        &            73.5 &                 88.3 &            81.0 \\
\hline
ALL (UniTransf)      &            90.4 &                 88.5 &            80.9 \\
ALL (BiTransf)       &            94.1 &                 94.1 &             - \\
\bottomrule
\end{tabular}
\caption{\small Cross-corpora generalization accuracy using TransfBig generator and UniTransf energy function (except for the last
row which used a bidirectional transformer).
Each row specifies the corpora used at training time, $C_{\text{train}}$. Each column shows the corpus used at test time, $C_{\text{test}}$.
\label{tab:cross-corpus}}
\end{table*}

\subsection{Generalization in the Wild}
In Table~\ref{tab:cross_architecture_huge} we test the generalization of the energy functions to GPT-2
generators~\citep{gpt2}\footnote{
We use pytorch versions of officially released checkpoints for small (137M params) and medium (380M params)
architectures from HuggingFace repository at \url{https://github.com/huggingface/pytorch-transformers}.}
that were trained on a completely different dataset, namely WebText~\citep{gpt2} a dataset of 8 million web pages.
This is an instance of unseen generalization since $C_{\text{train}} \neq C_{\text{test}}$, and
$A_{\text{train}} \neq A_{\text{test}}$.
We also consider generations from TransfHuge (last row) whose configuration is similar to the
unreleased biggest GPT2 model with 1.4 billion parameters, 7 times bigger than TransfBig, the generator used at training time.

Expectedly as the generator gets bigger the discrimination tasks gets harder.
When the energy function is confronted with generations from the GPT2 small model, which is
smaller than the training generator, the accuracy is close to the in-domain setting, however.
We notice that a bigger BiTransf model does not demonstrate significant advantage in this setting
and both discriminators achieve higher absolute accuracy on WebText with smaller generators.
That suggests that a big enough energy model trained with
a single big generator can efficiently discriminate a {\em block-box} generator. Of course, accuracy decreases
as the black-box generator is made bigger (GPT-2 medium).

\begin{table}[t]
  \center
\small
\begin{tabular}{lr|cc|cc}
\toprule
\multicolumn{2}{r|}{ Energy function  $\rightarrow$ }   & \multicolumn{2}{c|}{UniTransf trained on ALL} & \multicolumn{2}{c}{BiTransf trained on ALL} \\
 \multicolumn{2}{r|}{Test domain $\rightarrow$ }        & CCNews& WebCorpus                             & CCNews & WebCorpus                          \\
Generator model $\downarrow$  & M params $\downarrow$   & &                                           & &                                           \\
\midrule
GPT2 small (WebText)              &  137                & -    & 82.2                                   & -    & 86.0                                 \\
GPT2 medium (WebText)            &  380                 & -    & 73.8                                   & -    & 73.3                                 \\
\midrule
TransfHuge (CC-News)             & 1427                 & 65.4 & -                                      & 64.7 & -                                     \\
\bottomrule
\end{tabular}
\caption{\small
Generalization of BiTransf and UniTransf energy function to state of the art generators.
The energy functions (discriminators) are trained on the concatenation of all three corpora (same as in table~\ref{tab:cross-corpus}  ALL rows).
Test time negatives are generated by models specified in the rows, with their training set in parenthesis and model size in millions of parameters.  Note that both the training corpus and GPT2  generator are ``unseen'' by the energy function during training. \label{tab:cross_architecture_huge}}
\end{table}

Finally, we investigate generalization of the energy function to a new domain, such as samples from the dataset of
GPT-2 generations~\citep{gpt2gen}.
For each model the dataset has a 250k generated texts with either top-k sampling or random sampling.
Also, the dataset provides samples from the WebText corpus that was used to train the generator models, and
that we use to discriminate against.

In Table~\ref{tab:unconditinal} we report results of the BiTransf energy function compared to the TF-IDF baseline
provided with the dataset. We consider three cases.
In the in-domain setting, we finetune the energy function on the train set of each of the datasets, following the same protocol
used by the provided TF-IDF baseline.
In cross-architecture mode, we finetune only on the generations from the small GPT2 model (both top-k and random sampling),
and apply the model to the other datasets.
To adapt our models to this task we split the text segments into sets of intersecting blocks of 160 tokens.
During training we treat all blocks in a set as either positives or negatives.
During evaluation we take the mean prediction over all blocks in a segment as a prediction for the whole segment.

Finally, in the {\em wild} setting we explore generalization to a black-box generator. Similarly to WebCorpus column in table~\ref{tab:cross_architecture_huge}, our discriminator is trained on concatenation of CCNews, Books, and Wiki and tested on WebCorpus.
As we perform discrimination in unconditional setting, we train our energy model on all possible prefixes including an empty prefix.
Evaluation is performed in the same manner as for the in-domain and cross-architecture modes.

Unsurprisingly, in-domain BiTransf beats TF-IDF baseline getting almost 100\% across the board. 
However in cross-architecture mode, we can outperform the TF-IDF baseline only when the generator is less than
three times bigger than what was used at training time.
In the wild setting our discriminator works much better than a random predictor, yet it lags behind the simple (in-domain) linear baseline.
That suggests that matching the domain of the training set is more important than model complexity.

Interestingly, our energy function was trained using a fixed length input with a prefix. These generalization
results are significantly higher than the in-domain experiment of Table~\ref{tab:in_domain} because the unconditional task
is significantly easier, a topic further discussed next.

\begin{table}
\center
  \begin{tabular}{l||c|ccc}
    \multicolumn{1}{r||}{Energy Function $\rightarrow$ } & TF-IDF$^*$ & \multicolumn{3}{c}{BiTransf} \\
    \multicolumn{1}{r||}{Test setting $\rightarrow$ } & in-domain & in-domain & cross-architecture & wild  \\
    \midrule
 Small (137) top-k      & 96.79  & 99.09  &  -      &  93.25 \\
 Small (137) random    & 88.29  & 99.80  &  -       &  66.04 \\
 Med   (380) top-k      & 95.22  & 98.07  &  97.37  &  88.19 \\
 Med   (380) random    & 88.94  & 99.43  &  97.35   &  55.06 \\
 Big   (762) top-k      & 94.43  & 96.50  &  93.58  &  83.88 \\
 Big   (762)  random    & 77.16  & 99.42  &  95.96  &  64.03 \\
 Huge  (1542) top-k      & 94.43  & 95.01  &  90.17 &  79.18 \\
 Huge  (1542) random    & 77.31  & 99.00  &  91.76  &  61.29 \\
\bottomrule
  \end{tabular}
  \caption{\small
    \label{tab:unconditinal}
    Generalization of the energy function to {\em unconditional} generation from various GPT2 models (model size in
parantheses, followed by sampling method used). Each row contains the accuracy on the corresponding test set.
TF-IDF results are taken from~\citet{gpt2gen}. }
\end{table}

\subsection{Ablation Study} \label{sec:ablation}
\begin{figure}
\begin{floatrow}\CenterFloatBoxes
\ffigbox{%
\includegraphics[width=1.0\linewidth]{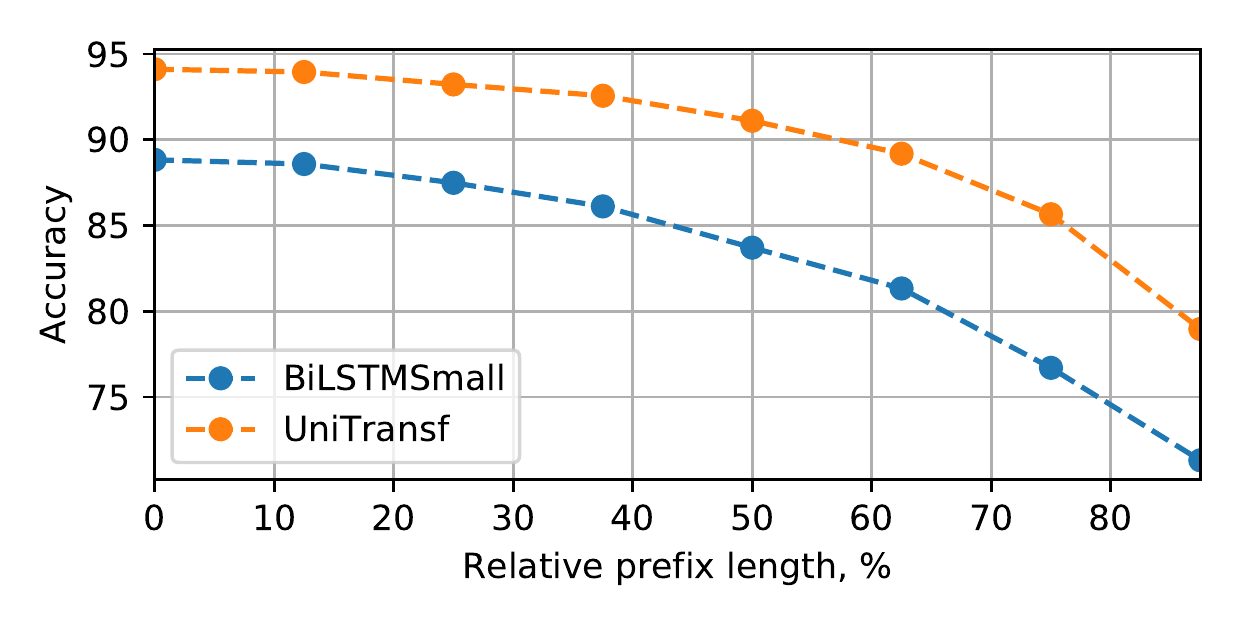}

}{%
\caption{\small
Discrimination accuracy as a function of the ratio between the prefix length and the total length of the sequence
on the Wikitext dataset.
}
\label{fig:accuracy_length}
}
\capbtabbox{%
\begin{tabular}{lc}
\toprule
&  Accuracy \\
\midrule
1 random negative
&                   82.9 \\
3 random negatives
&                   84.2 \\
worst negative out of 3
&                   84.6 \\
\bottomrule
\end{tabular}
}{%
\caption{\small Effect of different strategies to mine negatives using
TransfBig generator and BiLSTMSmall energy function on Book Corpus.
}
\label{tab:ablationloss}
}
\end{floatrow}
\end{figure}

First, we investigate the dependency between performance of the energy functions and length of the prefix.
We trained BiLSTMSmall and UniTransf models on examples with varying prefix length from the Wikitext corpus,
and computed the accuracy for each prefix length independently.
Figure~\ref{fig:accuracy_length} shows that as the prefix length increases (and the generation gets shorter),
the discrimination task gets harder and the difference between the models more prominent.
The unconditional case, i.e. zero prefix length, is the easiest, while prefixes of length 120 and 140 that are the main
experimental setup in this work, are the hardest.

Finally, in Table~\ref{tab:ablationloss} we study the impact of the number of negatives and using the most offending
negative in the loss of Eq.~\eqref{eq:bceloss}. Using more negatives and harder negatives improves
accuracy.

\subsection{Stability to Other Negative Distributions}
\begin{figure}[!t]
\centering
\includegraphics[width=\linewidth]{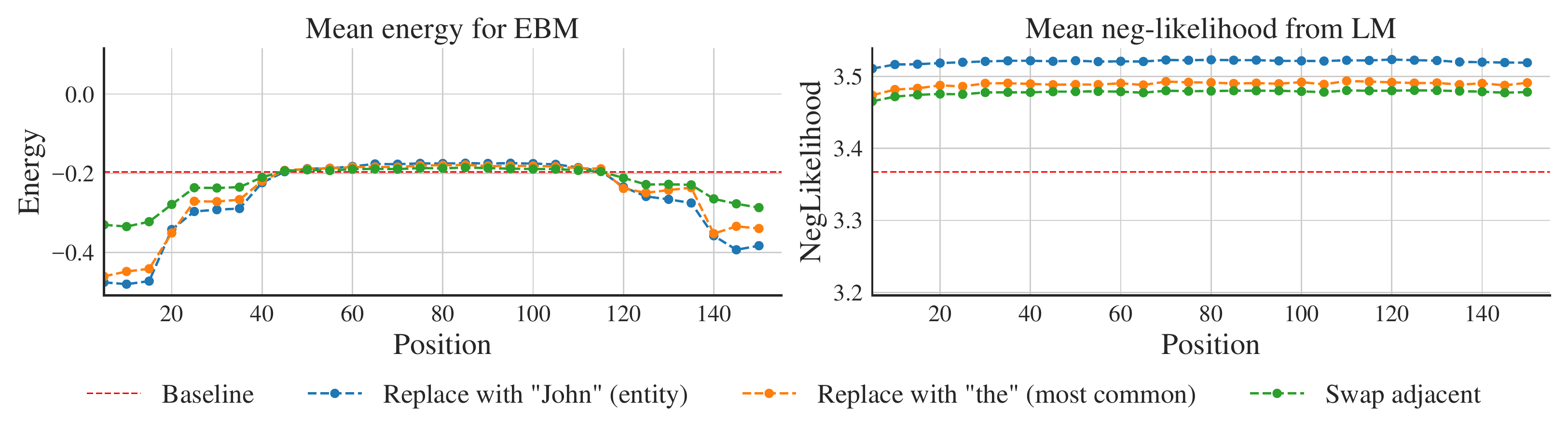}
\caption{\small
Effect of applying various perturbations (word replacement and swap of adjacent words) to ground-truth sequences
at different positions in terms of energy function and generator negative log-likelihood
(averaged over the whole test set of Wikitext).
The energy is only affected by corruptions at either end of the sequence.
These out-of-domain corruptions invariably decrease the energy.
  However, all perturbations increase the negative log-likelihood of the sequence.}
\label{fig:perturbation}
\end{figure}
In the previous sections we have seen that the energy function is less robust to negatives generated
from a model trained on a different corpus.
However, even in that case, a negative is still a sample from an auto-regressive neural network.
In Appendix~\ref{sec:exploring}, we show examples where changing a few entities can cause large
jumps in the energy (from negative to positive or vice versa), and so fool the EBM.
More generally, we see that the energy function is not robust to truly out-of-domain samples.
For example, the energy will score blocks of randomly generated text lower than real text.

These behaviors are evidence that the energy functions have learned the regularities of {\em generated} text,
as opposed to learning the regularities of real text. We surmise that it does so because modeling the latter would be much
more difficult than the former. By modeling generated text,
the energy function assigns low score to anything that is not generated by its training generator.

 While not surprising, this might be considered a liability of such energy functions.
However, as a model of text, the energy functions should be considered as working on the {\em residuals} of the
language models used to generate negatives.
   For the examples in Appendix~\ref{sec:exploring}, the language model records a large {\em decrease}
in likelihood after the change in entity; and language models of course give much lower likelihood to random text
than gold or generated text. Therefore, the energy function needs not to be accurate on examples that are already
very unlikely according to these language models.

In Figure~\ref{fig:perturbation} we show the average effects of applying various perturbations to sequences from
Wikitext103  on an in-domain energy and language model at each location (from 1 to 160) in the sequence.
We see that for all perturbations, the energy decreases its value, but the language model increases its negative log likelihood.
 We also see that the energy function is more sensitive to the ends of the text, which is where the negatives were different
from real text at training time.

\section{Final Remarks}
The EBM framework could potentially unlock more expressive models of text, as they are not limited to scoring a
single word at a time as current locally normalized auto-regressive models do.  Unfortunately, training EBMs is challenging because generating negatives using the energy function itself is still an open research problem, and does not scale well in practice. In this work, we propose a simple solution, which is to leverage generations produced by
pre-trained language models as negative samples.

As a preliminary yet necessary step in this direction we have investigated the generalization ability of such EBMs.
We found that EBMs, when trained on large datasets, achieve  good generalization. For instance, they
behave nicely when tested with negatives produced by generators that have rather different architectures.
The generalization is less good when generators are trained on other corpora, but EBMs re-gain robustness once we train
them on even bigger composite datasets.

In the future, we can improve EBMs for text  by simply making their architectures bigger and
increasing the diversity and size of their training datasets. Of course, further scaling up of EBMs will pose
formidable engineering challenges.

On the application side, a natural application of the current formulation of EBMs is real/fake text discrimination.
We believe that  this is important application in its own right, and that EBMs can be very powerful, as demonstrated
by their superior performance compared to discriminating using the original language model log-likelihood.

We additionally hope to broaden the scope of applications of EBMs for text, including learning generic representations
of text and using EBMs to improve text generation.

\bibliographystyle{iclr2020_conference}
\bibliography{refs}

\newpage
\appendix
\section{Corpora sizes}
\begin{table}[!h]
\center
\begin{tabular}{l|rrr}
\bf Dataset & Train & Valid & Test \\
\toprule
Books &  690 & 7.3 & 8.0 \\
CCNews & 21718 & 1.0 & 3.4 \\
Wikitext &   113 & 0.2& 0.3\\
\end{tabular}
\caption{\small Number of BPE tokens in millions for each dataset.}
\label{tab:datasets}
\end{table}

\section{Model Sizes} \label{sec:sizes}
\begin{table}[!h]
\center
\begin{tabular}{l|cccc|cccc}
\toprule
& \multicolumn{8}{c}{\bf Generators} \\
& \multicolumn{4}{c}{} & \multicolumn{4}{c}{Pre-trained GPT2} \\
& Conv & TransfSmall & TransfBig & TransfHuge & small & med & large & huge \\
\midrule
embed. & 13 & 26  &  51 & 77   & 39  &  52 & - & - \\
others & 164 & 19 & 151 & 1360 & 97  & 327 & - & - \\
total & 176 & 45  & 203 & 1437 & 137\footnote{We use models from HuggingFace repository (\url{https://github.com/huggingface/pytorch-transformers}) and report here the sizes of these models as they were used to generate data for table~\ref{tab:cross_architecture_huge}. Note that the OpenAI GPT2 repository (\url{https://github.com/openai/gpt-2}) defines models sizes as 124M and 355M for small and medium model correspondingly. \label{fn:repeat}} & 380\footref{fn:repeat}  & 762\footnote{As reported in~\cite{gpt2}.\label{fn:gpt2_paper}} & 1542\footref{fn:gpt2_paper} \\
\bottomrule
\end{tabular}
\caption{\small Number of parameters (in millions) for the generator language models.
The computational cost is directly related to the number of parameters in other layers than the input embedding layer (second row).}
\label{tbl:generation_models_size}
\end{table}

\begin{table}[!h]
\center
\begin{tabular}{l|ccccc}
\toprule
&  \multicolumn{5}{c}{\bf EBM Functions} \\
&  Linear & BiLSTM & BiLSTM Big &  UniTransf & BiTransf \\
\midrule
embed. &        0.1  & 26 & 39  &  51  & 51 \\
others &   0         & 23 & 90  &  151 & 304 \\
total &          0.1 & 49 & 129 &  203 & 355 \\
\bottomrule
\end{tabular}
\caption{\small Number of parameters in millions for the scoring functions.
The computational cost is directly related to the number of parameters in other layers than the input embedding layer (second row).
\label{tbl:scoring_models_size}}
\end{table}


\section{Ranking Loss} \label{sec:ranking}
The (per-sample) ranking loss is:
\begin{equation}
\mathcal{L}_{\mbox{R}} = \max \left(0, 1 + E(x^+ | c;\theta) - E(x^-_i | c; \theta)\right).
\label{eq:rankloss}
\end{equation}
In this case we also refer to the negative energy as the model {\em score}.
The ranking loss makes the energy values {\em local}, as the loss takes as input the difference of energies for
a pairs of positive and negative that share the {\em same} context. Instead, the binary cross entropy loss of
Eq.~\ref{eq:bceloss} encourages
a more {\em global} and absolute scoring as the loss forces all positive examples to have negative energy, and all
negative samples to have positive energy, regardless of the context. Therefore, the binary cross entropy loss is
perhaps more interpretable as it is not context dependent, but the task is also harder to learn. Empirically, we found
similar findings with both losses.

When the energy function is trained using the ranking loss of eq.~\ref{eq:rankloss}, we evaluate the model using
{\em precision at 1} (P@1), which is the ratio between the number of times the ground truth sequence scores the lowest
over its set of negatives, averaged over the number of sequences in the test set.

\section{Hyper-parameter Setting}
All models are implemented using the PyTorch framework~\citep{paszke2017automatic} and are optimized using Adam~\citep{kingma2014adam}.

To train our biggest models (UniTransf and BiTransf) we used 8 machines each with 8 GPUs in synchronous mode using data parallelism.
The resulting large batch size speeds up training when combined with float16 reduced precision and cosine scheduling of the learning rate
without any restarts~\citep{loshchilov2016sgdr}, i.e. we decay the learning rate to zero over the course of ``max steps'' updates and then stop training.
Using these methods, we reduced training time by five times compared to a single node training.
For all other configurations we used a single node with up to 8 GPUs and inverse square root decay.

\begin{table}[!h]
\center
\begin{tabular}{l|cccccc}
\toprule
& max lr & bsz (per GPU) & GPUs & fp16 & warmup steps & max steps\\
\midrule
Linear              & 0.01  & 1024 & 1   & + & 1000 & -      \\
BiLSTM              & 0.0002 & 128 & 8   & + & 1000 & -      \\
UniTransf           & 0.0003 & 32  & 64  & + & 2000 & 180000 \\
BiTransf            & 0.00005& 20  & 192 & + & 2000 & 180000 \\
\bottomrule
\end{tabular}
\caption{\small Hyper-parameter values used in our scoring functions.}
\label{tbl:hyperparams}
\end{table}

\newpage
\section{Score Distributions}
\begin{figure}[!h]
\centering
\includegraphics[width=.99\linewidth]{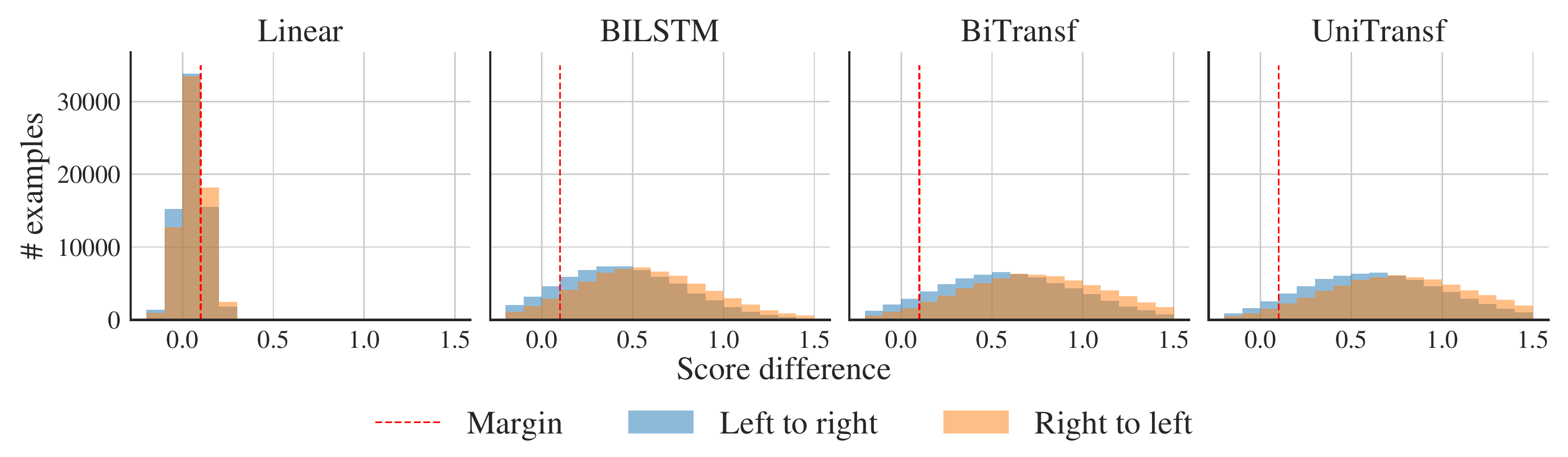}
\vspace{-2mm}
\caption{\small Distributions of score (negative energy) differences between pairs of ground truth completions and generated
ones for different scoring models.
  We show results for two generations (left to right and right to left) from Wikitext dataset.
  In both cases we generate 40 tokens.  Examples on the right of the red line ($margin=0.1$) have zero rankiing loss.}
\label{fig:margins}
\end{figure}

\section{\label{sec:exploring}Perturbing the Energy Function}
In this section we show that we can change a few words to make a negative example become a ``positive'' one
as judged by the energy function, and vice versa, by using gradient information.

Below here, we show an example of a ground truth sentence from the Wikitext dataset.

\begin{framed}
 <EOS> =Robert Boulter= <EOS>  <EOS> Robert Boulter is an English film, television and theatre actor. He had a guest-starring role on the television series The Bill in 2000. This was followed by a starring role in the play Herons written by Simon Stephens, which was performed in 2001 at the Royal Court Theatre. He had a guest role in the television series Judge John Deed in 2002. In 2004 Boulter landed a role as "Craig" in the episode "Teddy's Story" of the television series The Long Firm; he starred alongside actors Mark Strong and\generategap{ Derek Jacobi. He was cast in the 2005 theatre productions of the Philip Ridley play Mercury Fur, which was performed at the Drum Theatre in Plymouth and the Menier Chocolate Factory in London. He was}
\end{framed}

Here the block has 160 BPE tokens, where the first 120 tokens (black font) are used as context and the remaining
40 are the ground truth completion.
Next, we use a language model to generate 10 negatives:

\begin{framed}
\paragraph{Negative 1} <EOS> =Robert Boulter= <EOS>  <EOS> Robert Boulter is an English film, television and theatre actor. He had a guest-starring role on the television series The Bill in 2000. This was followed by a starring role in the play Herons written by Simon Stephens, which was performed in 2001 at the Royal Court Theatre. He had a guest role in the television series Judge John Deed in 2002. In 2004 Boulter landed a role as "Craig" in the episode "Teddy's Story" of the television series The Long Firm; he starred alongside actors Mark Strong and\generategap{ Chris Elliott in 2006 as the character. Boulter has appeared in various television specials dealing with the series since its inception. <EOS> After graduating with a degree in drama, Boulter worked as a}

\paragraph{Negative 2} <EOS> =Robert Boulter= <EOS>  <EOS> Robert Boulter is an English film, television and theatre actor. He had a guest-starring role on the television series The Bill in 2000. This was followed by a starring role in the play Herons written by Simon Stephens, which was performed in 2001 at the Royal Court Theatre. He had a guest role in the television series Judge John Deed in 2002. In 2004 Boulter landed a role as "Craig" in the episode "Teddy's Story" of the television series The Long Firm; he starred alongside actors Mark Strong and\generategap{ Stephen Fry in the episode "You're All Alone" and in the episode "The Longest Day". <EOS> He auditioned for the role in the series in 2003 but was not cast. In 2005}

$\vdots$

\paragraph{Negative 10} <EOS> =Robert Boulter= <EOS>  <EOS> Robert Boulter is an English film, television and theatre actor. He had a guest-starring role on the television series The Bill in 2000. This was followed by a starring role in the play Herons written by Simon Stephens, which was performed in 2001 at the Royal Court Theatre. He had a guest role in the television series Judge John Deed in 2002. In 2004 Boulter landed a role as "Craig" in the episode "Teddy's Story" of the television series The Long Firm; he starred alongside actors Mark Strong and\generategap{ Ian Somerhalder on the BBC series Top Gear; this was followed up in 2007 by a role in the BBC science-fiction series Doctor Who. In 2008 Boulter appeared in the BBC}
\end{framed}

\begin{figure}[!h]
\centering
\includegraphics[width=.7\linewidth]{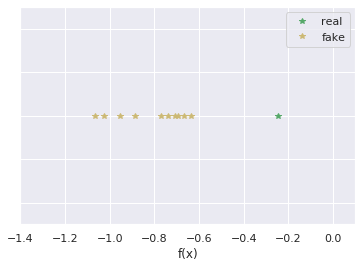}
\caption{\small Real and fake (negatives generated from TransfBig language model) completions as scored by the learned energy function.
The energy function is able to separate them well. These scores are calcuated based on the single example
reported in the main text of \textsection\ref{sec:exploring}. $f(x)$ is the negative energy.}
\label{fig:margin_single}
\end{figure}

In this example, using the big transformer model, UniTransf, as the energy function,
we are able to separate real from fake examples
as shown (Figure~\ref{fig:margin_single}). We want to perturb these negatives to violate the margin.
To do so, we make use of the gradient information from the energy function $\nabla_{x} E_\theta (x)$
and use a first order Taylor expansion to
approximate the effect of a token replacement (we abuse our notations and use $x$ to denote embeddings in this analysis).
Given the original sample $x$, we change one word $x_i$ to $x_i'$ to arrive at $x'$. The score of $x'$ is approximately:

\begin{equation*}
E_\theta(x) + \nabla_{x_i} E_\theta(x) \cdot (x_i'-x_i)
\end{equation*}

Using this approximation, we can search for those token replacements that increase/decrease the energy the most.
We can easily change a negative sample to a positive one by replacing the 5 words highlighted below.
 In paratheses, we report both score and language model perplexity.

\begin{framed}
\paragraph{Original negative (score -0.77, PPL 20.77)}  <EOS> =Robert Boulter= <EOS>  <EOS> Robert Boulter is an English film, television and theatre actor. He had a guest-starring role on the television series The Bill in 2000. This was followed by a starring role in the play Herons written by Simon Stephens, which was performed in 2001 at the Royal Court Theatre. He had a guest role in the television series Judge John Deed in 2002. In 2004 Boulter landed a role as "Craig" in the episode "Teddy's Story" of the television series The Long Firm; he starred alongside actors Mark Strong and\markgap{ Chris}\markgap{ Elliott} in 2006 as the character. Boulter has appeared in various television specials\markgap{ dealing} with the series since its inception. <EOS> After graduating with a degree in\markgap{ drama}, Boulter worked as a

\paragraph{Perturbed negative (score 0.00, PPL 117.30)} <EOS> =Robert Boulter= <EOS>  <EOS> Robert Boulter is an English film, television and theatre actor. He had a guest-starring role on the television series The Bill in 2000. This was followed by a starring role in the play Herons written by Simon Stephens, which was performed in 2001 at the Royal Court Theatre. He had a guest role in the television series Judge John Deed in 2002. In 2004 Boulter landed a role as "Craig" in the episode "Teddy's Story" of the television series The Long Firm; he starred alongside actors Mark Strong and\markgapsingle{ Gor}{-0.0.64}{28.97}\markgapsingle{ Trem}{-0.56}{38.86} in 2006 as the character. Boulter has appeared in various television specials\markgapsingle{ relates}{-0.77}{24.60} with the series since its inception. <EOS> After\markgapsingle{Health}{-0.35}{39.52} with a degree in\markgapsingle{edited}{-0.49}{27.45}, Boulter worked as a
\end{framed}

In the above example, we also show the (score, PPL) for replacing a single token in the subscripts. Similarly, we can replace a few words and make a positive sample become negative.

\begin{framed}
\paragraph{Original positive (score -0.25, PPL 77.68)}  <EOS> =Robert Boulter= <EOS>  <EOS> Robert Boulter is an English film, television and theatre actor. He had a guest-starring role on the television series The Bill in 2000. This was followed by a starring role in the play Herons written by Simon Stephens, which was performed in 2001 at the Royal Court Theatre. He had a guest role in the television series Judge John Deed in 2002. In 2004 Boulter landed a role as "Craig" in the episode "Teddy's Story" of the television series The Long Firm; he starred alongside actors Mark Strong and\markgap{ Derek} Jacobi. He was cast in the 2005 theatre productions of the Philip Ridley play Mercury Fur, which was performed at the\markgap{ Drum} Theatre in\markgap{ Plymouth} and the\markgap{ Men}ier\markgap{ Chocolate} Factory in London. He was

\paragraph{Perturbed positive (score -0.78, PPL 142.85)}  <EOS> =Robert Boulter= <EOS>  <EOS> Robert Boulter is an English film, television and theatre actor. He had a guest-starring role on the television series The Bill in 2000. This was followed by a starring role in the play Herons written by Simon Stephens, which was performed in 2001 at the Royal Court Theatre. He had a guest role in the television series Judge John Deed in 2002. In 2004 Boulter landed a role as "Craig" in the episode "Teddy's Story" of the television series The Long Firm; he starred alongside actors Mark Strong and\markgapsingle{connected}{-0.30}{118.30} Jacobi. He was cast in the 2005 theatre productions of the Philip Ridley play Mercury Fur, which was performed at the\markgapsingle{ C}{-0.28}{75.36} Theatre in\markgapsingle{ London}{-0.47}{62.29} and the\markgapsingle{ Vaughan}{-0.40}{93.77}ier\markgapsingle{cerning}{-0.32}{100.71} Factory in London. He was
\end{framed}

\begin{figure}[!h]
\centering
\includegraphics[width=.7\linewidth]{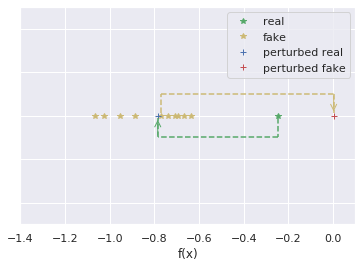}
\caption{\small By changing a few words we can make a negative sample become real as scored by the (negative)
ennergy function, and vice versa.}
\label{fig:margin_edited}
\end{figure}

As shown in Figure~\ref{fig:margin_edited}, we can easily ``fool'' the discriminator by editing a few words.
However, these edited sentences have a very low probability (high PPL) under the generator we used.
This explains why the discriminator gets fooled, because it has never seen such negatives during training.

\end{document}